\documentclass[11pt]{article}
\usepackage{acl2015}
\usepackage{times}
\usepackage{url}
\usepackage{amsmath,amsfonts,amssymb}
\usepackage{latexsym}
\usepackage{caption}
\usepackage{subcaption}
\usepackage{algorithm}
\usepackage{algorithmic}
\usepackage{graphicx}
\usepackage{color}

\usepackage{bbm}
\usepackage{dsfont}
\usepackage{array}

\newcommand{\comment}[1]{}

\DeclareMathOperator*{\argmax}{arg\,max}
\DeclareMathOperator*{\argmin}{arg\,min}

\title{Learning Dynamic Feature Selection for Fast Sequential Prediction}

\author{Emma Strubell \qquad Luke Vilnis \qquad Kate Silverstein \qquad Andrew McCallum\\
  College of Information and Computer Sciences \\
  University of Massachusetts Amherst \\
  Amherst, MA, 01003, USA \\
  {\tt \{strubell, luke, ksilvers, mccallum\}@cs.umass.edu}}

\date{}

\begin{document}

\maketitle

\begin{abstract}
We present paired learning and inference algorithms for significantly
reducing computation and increasing speed of the vector dot products
in the classifiers that are at the heart of many NLP components.  This
is accomplished by partitioning the features into a sequence of
templates which are ordered such that high confidence can often be
reached using only a small fraction of all features.  Parameter
estimation is arranged to maximize accuracy and early confidence in
this sequence.  Our approach is simpler and better suited to NLP than
other related cascade methods.  We present experiments in
left-to-right part-of-speech tagging, named entity recognition, and transition-based dependency parsing. On the typical benchmarking datasets we can preserve POS tagging accuracy above 97\% and parsing LAS above 88.5\% both with over a five-fold reduction in run-time, and NER F1 above 88 with more than 2x increase in speed.
\end{abstract}

\section{Introduction}

Many NLP tasks such as part-of-speech tagging, parsing and named entity recognition have become sufficiently accurate that they are no longer solely an object of research, but are also widely deployed in production systems.  These systems can be run on billions of documents, making the efficiency of inference a significant concern---impacting not only wall-clock running time but also computer hardware budgets and the carbon footprint of data centers.

This paper describes a paired learning and inference approach for
significantly reducing computation and increasing speed while
preserving accuracy in the linear classifiers typically used in many
NLP tasks. The heart of the prediction computation in these models is
a dot-product between a dense parameter vector and a sparse feature
vector. The bottleneck in these models is then often a combination of
feature extraction and numerical operations, each of which scale
linearly in the size of the feature vector. Feature extraction can be
even more expensive than the dot products, involving, for example,
walking sub-graphs, lexicon lookup, string concatenation and string
hashing.  \comment{Predicting with these models then involves on
  average 50 multiplications and 50 additions (with additional
  exponentiation and normalization when computing expectations and
  likelihoods in probabilistic models).  In addition to these
  numerical operations, the features themselves must be produced from
  the relevant context, involving, for example, walking sub-graphs,
  lexicon lookup, string concatenation and string hashing.}  We note,
however, that in many cases not all of these features are necessary
for accurate prediction.  For example, in part-of-speech tagging if we
see the word ``the,'' there is no need to perform a large dot product
or many string operations; we can accurately label the word a {\sc
  Determiner} using the word identity feature alone. In other cases
two features are sufficient: when we see the word ``hits'' preceded by
a {\sc Cardinal} (e.g. ``two hits'') we can be confident that it is a
{\sc Noun}.

We present a simple yet novel approach to improve processing speed by
dynamically determining on a per-instance basis how many features are
necessary for a high-confidence prediction. Our features are divided
into a set of \emph{feature templates}, such as {\sf\small
  current-token} or {\sf\small previous-tag} in the case of POS
tagging. At training time, we determine an ordering on the templates
such that we can approximate model scores at test time by
incrementally calculating the dot product in template ordering. We
then use a running confidence estimate for the label prediction to
determine how many terms of the sum to compute for a given instance,
and predict once confidence reaches a certain threshold.

In similar work, cascades of increasingly complex and high-recall models have been used for both structured and unstructured prediction. \newcite{viola2001rapid} use a cascade of boosted models to perform face detection. \newcite{weiss-taskar-10} add increasingly higher-order dependencies to a graphical model while filtering the output domain to maintain tractable inference. While most traditional cascades pass instances down to layers with increasingly higher recall, we use a single model and accumulate the scores from each additional template until a label is predicted with sufficient confidence, in a stagewise approximation of the full model score. Our technique applies to any linear classifier-based model over feature templates without changing the model structure or decreasing prediction speed. 

Most similarly to our work, \newcite{weiss-taskar-13} improve performance for several structured vision tasks by dynamically selecting features at runtime. However, they use a reinforcement learning approach whose computational tradeoffs are better suited to vision problems with expensive features. Obtaining a speedup on tasks with comparatively cheap features, such as part-of-speech tagging or transition-based parsing, requires an approach with less overhead. In fact, the most attractive aspect of our approach is that it speeds up methods that are already among the fastest in NLP.

We apply our method to left-to-right part-of-speech tagging in which we achieve accuracy above 97\% on the Penn Treebank WSJ corpus while running more than five times faster than our 97.2\% baseline. We also achieve a five-fold increase in transition-based dependency parsing on the WSJ corpus while achieving an LAS just 1.5\% lower than our 90.3\% baseline. Named entity recognition also shows significant speed increases. We further demonstrate that our method can be tuned for $2.5-3.5$x multiplicative speedups with nearly no loss in accuracy. 

\section{Classification and Structured Prediction}
\label{sec:sp}

Our algorithm speeds up prediction for multiclass
classification problems where the label set can be tractably
enumerated and scored, and the per-class scores of input features
decompose as a sum over multiple feature templates. 
Frequently, classification problems in NLP
are solved through the use of linear classifiers, which compute scores
for input-label pairs using a dot product. These meet our additive
scoring criteria, and our acceleration methods are directly applicable.

However, in this work we are interested in speeding up \emph{structured prediction} problems,
specifically part-of-speech (POS) tagging and dependency parsing. We apply
our classification algorithms to these problems by reducing them to
\emph{sequential prediction} \cite{daume2009search}. For POS tagging, 
we describe a sentence's part of speech annotation by
the left-to-right sequence of tagging decisions for individual tokens
\cite{Gimenez-Marquez-2004}. Similarly, we implement our parser with a classifier that generates a sequence
of shift-reduce parsing transitions \cite{nivre2009non}. 

The use of sequential prediction to solve these problems and others has a long history in practice as well as theory. Searn \cite{daume2009search} and DAgger \cite{RossGB11} are two popular principled frameworks for reducing sequential prediction to classification by learning a classifier on additional synthetic training data. However, as we do in our experiments, practitioners often see good results by training on the gold standard labels with an off-the-shelf classification algorithm, as though classifying IID data \cite{bengtson2008understanding,choi-palmer-12}. 

Classifier-based approaches to structured prediction are faster than dynamic programming since they consider only a subset of candidate output structures in a greedy manner.
For example, the Stanford CoreNLP
classifier-based part-of-speech tagger provides a 6.5x speed advantage
over their dynamic programming-based model, with little reduction in
accuracy. Because our methods are designed for the greedy sequential prediction regime, we can provide further speed increases to the fastest inference methods in NLP.

\section{Linear models}
\label{linear-model-section}

Our base classifier for sequential prediction tasks will be a
\emph{linear model}. Given an input $x \in \mathcal{X}$, a set of
labels $\mathcal{Y}$, a feature map $\Phi(x,y)$, and a weight vector
$\mathbf{w}$, a linear model predicts the highest-scoring label 
\begin{align}
\label{eq:linear}
y^* = \argmax_{y \in \mathcal{Y}} ~\mathbf{w} \cdot \Phi(x, y).
\end{align}
The parameter $\mathbf{w}$ is usually learned by minimizing a
regularized ($R$) sum of loss functions ($\ell$) over the training
examples indexed by $i$
\begin{align*}
\mathbf{w}^* = \argmin_{\mathbf{w}} \sum_{i} \ell(x_i, y_i, \mathbf{w}) + R(\mathbf{w}).
\end{align*}

In this paper, we partition the features into a set of \emph{feature templates}, so that the weights, feature function, and dot product factor as
\begin{align}
\label{eq:templates}
\mathbf{w} \cdot \Phi(x, y) = \sum_j \mathbf{w}_j \cdot \Phi_j(x,y)
\end{align}
for some set of \emph{feature templates} $\{\Phi_j(x,y)\}$. 

Our goal is to approximate the dot products in \eqref{eq:linear}
sufficiently for purposes of prediction, while using as few terms of
the sum in \eqref{eq:templates} as possible.

\section{Method}
\label{sec:alg}

We accomplish this goal by developing paired learning and
inference procedures for feature-templated classifiers that optimize
both accuracy and inference speed, using a process of \emph{dynamic
  feature selection}. Since many decisions are easy to make in the
presence of strongly predictive features, we would like our model to
use fewer templates when it is more confident. For a fixed, learned
ordering of feature templates, we build up a vector of class scores
incrementally over each prefix of the sequence of templates, which we
call the \emph{prefix scores}. Once we reach a stopping criterion
based on class confidence (margin), we stop computing prefix 
scores, and predict the current highest scoring class. Our aim is to train
each prefix to be as good a classifier as possible without the
following templates, minimizing the number of templates
needed for accurate predictions.

Given this method for performing fast inference on an ordered set of
feature templates, it remains to choose the ordering. In Section \ref{ordering-section},
we develop several methods for picking template orderings, based on
ideas from group sparsity \cite{yuan2006model,swirszcz2009grouped}, and other techniques for feature subset-selection \cite{kohavi1997wrappers}.

\subsection{Definitions}

Given a model that computes scores additively over template-specific scoring functions as in \eqref{eq:templates}, parameters $\mathbf{w}$, and an observation $x \in {X}$, we can define the $i$'th \emph{prefix score} for label $y \in \mathcal{Y}$ as:
\begin{align*}
P_{i,y}(x, \mathbf{w}) = \sum_{j=1}^i \mathbf{w}_j \cdot \Phi_j(x,y),
\end{align*}
or $P_{i,y}$ when the choice of observations and weights is clear from context. Abusing notation we also refer to the vector containing all $i$'th prefix scores for observation $x$ associated to each label in $\mathcal{Y}$ as $P_i(x, \mathbf{w})$, or $P_i$ when this is unambiguous.

Given a parameter $m > 0$, called the \emph{margin}, we define a function $h$ on prefix scores:
\begin{align*}
h(P_i, y) &= \max\{0,\max_{y'\neq y} P_{i,y'} - P_{i,y} + m\}
\end{align*}
This is the familiar structured hinge loss function as in structured support vector machines \cite{tsochantaridis2004support}, which has a minimum at $0$ if and only if class $y$ is ranked ahead of all other classes by at least $m$.

Using this notation, the condition that some label $y$ be ranked first by a margin can be written as $h(P_i,y) = 0$, and the condition that any class be ranked first by a margin can be written as $\max_{y'} h(P_i,y') = 0$.

\begin{algorithm}[tb]
  \caption{Inference}
  \label{alg:infer}
\begin{algorithmic}
   \STATE {\bfseries Input:} template parameters $\{\textbf{w}_i\}_{i=1}^k$, margin $m$
   \STATE and optional (for train time) true label $y$
   \STATE {\bfseries Initialize:} $i=1$
   \WHILE{$l > 0 \wedge i \leq k$}
   \STATE $l = \max_{y'} h(P_i,y')$ (test) or $h(P_i,y)$ (train)
   \STATE $i \leftarrow i + 1$
   \ENDWHILE
   \RETURN $\{P_j\}_{j=1}^i$ (train) or $\max_{y'} P_{i,y'}$ (test)
\end{algorithmic}
\end{algorithm}


\begin{algorithm}[tb]
  \caption{Parameter Learning}
  \label{alg:learn}
\begin{algorithmic}
   \STATE {\bfseries Input:} examples $\{(x_i,y_i)\}_i^N$, margin $m$
   \STATE {\bfseries Initialize:} parameters $\textbf{w}_0=0$, $i=1$
   \WHILE{$i \leq N$}
   \STATE $\text{prefixes}~\leftarrow~\text{Infer}(x_i,y_i,\textbf{w}_i,m)$
   \STATE $g_i\leftarrow\text{ComputeGradient}(\text{prefixes})$
   \STATE $\textbf{w}_{i+1}\leftarrow\text{UpdateParameters}(\textbf{w}_{i}, g_i)$
   \STATE $i \leftarrow i + 1$
   \ENDWHILE
   \RETURN $\textbf{w}_N$
\end{algorithmic}
\end{algorithm}


\subsection{Inference}

As described in Algorithm \ref{alg:infer}, at test time we compute prefixes until some label is ranked ahead of all other labels with a margin $m$, then predict with that label. At train time, we predict until the correct label is ranked ahead with margin $m$, and return the whole set of prefixes for use by the learning algorithm. If no prefix scores have a margin, then we predict with the final prefix score involving all the feature templates.

\subsection{Learning}
\label{learning-section}

We split learning into two subproblems: first, given an ordered sequence of feature templates and our inference procedure, we wish to learn parameters that optimize accuracy while using as few of those templates as possible. Second, given a method for training feature templated classifiers, we want to learn an ordering of templates that optimizes accuracy.

We wish to optimize several different objectives during learning: template parameters
should have strong predictive power on their own,
but also work well when combined with the scores from later templates. Additionally,
we want to encourage well-calibrated confidence scores that allow us to stop
prediction early without significant reduction in generalization ability.

\subsection{Learning the parameters}
\label{param-learning-section}

To learn parameters that encourage the use of few feature templates, we look at the model as outputting not a single prediction but a sequence of prefix predictions $\{P_i\}$. For each training example, each feature template receives a number of hinge-loss gradients equal to its distance from the index where the margin requirement is finally reached. This is equivalent to treating each prefix as its own model for which we have a hinge loss function, and learning all models simultaneously. Our high-level approach is described in Algorithm \ref{alg:learn}.

Concretely, for $k$ feature templates we optimize the following structured max-margin objective (with the dependence of $P$'s on $\mathbf{w}$ written explicitly where helpful):
\begin{align*}
\textbf{w}^* &= \argmin_{\textbf{w}} \sum_{(x,y)} \ell(x,y,\textbf{w})\\
\ell(x,y, \textbf{w}) &= \sum_{i=1}^{i^*_y} h(P_i(x, \textbf{w}), y)\\
i^*_y &= \min \{i\}^{k}_{i=1} \mathrm{\ \ \ s.t.\ \ \ } h(P_i,y) = 0
\end{align*}

The per-example gradient of this objective for weights $\textbf{w}_j$ corresponding to feature template $\Phi_j$ then corresponds to
\begin{align*}
\frac{\partial \ell}{\partial \mathbf{w}_j} \sum_{i=j}^{i^*_y} \Phi_j(x, y_{\text{loss}}(P_i,y)) - \Phi_j(x, y).
\end{align*}
where we define 
\begin{align*}
y_{\text{loss}}(P_i,y) &= \argmax_{y'} P_{i,y'} - m\cdot I(y' = y),
\end{align*}
where $I$ is an indicator function of the label $y$, used to define loss-augmented inference.

We add an $\ell_2$ regularization term to the objective, and tune the margin $m$ and the regularization strength to tradeoff between speed and accuracy. In our experiments, we used a development set to choose a regularizer and margin that reduced test-time speed as much as possible without decreasing accuracy. We then varied the margin for that same model at test time to achieve larger speed gains at the cost of accuracy. In all experiments, the margin with which the model was trained corresponds to the largest margin reported, i.e. that with the highest accuracy.

\subsection{Learning the template ordering}
\label{ordering-section}

We examine three approaches to learning the template ordering.

\subsubsection{Group Lasso and Group Orthogonal Matching Pursuit}

The Group Lasso regularizer \cite{yuan2006model} penalizes the sum of $\ell_2$-norms of weights of feature templates (different from what is commonly called ``$\ell_2$'' regularization, penalizing squared $\ell_2$ norms), $\sum_i c_i \|w_i\|_2$, where $c_i$ is a weight for each template. This regularizer encourages entire groups of weights to be set to $0$, whose templates can then be discarded from the model. By varying the strength of the regularizer, we can learn an ordering of the importance of each template for a given model. The included groups for a given regularization strength are nearly always subsets of one another (technical conditions for this to be true are given in \newcite{hastie2007forward}). The sequence of solutions for varied regularization strength is called the \emph{regularization path}, and by slight abuse of terminology we use this to refer to the induced template ordering.

An alternative and related approach to learning template orderings is based on the Group Orthogonal Matching Pursuit (GOMP) algorithm for generalized linear models \cite{swirszcz2009grouped,lozano2011group}, with a few modifications for the setting of high-dimensional, sparse NLP data (described in Appendix \ref{gomp-derivation}). Orthogonal matching pursuit algorithms are a set of stagewise feature selection techniques similar to forward stagewise regression \cite{hastie2007forward} and LARS \cite{efron2004least}. At each stage, GOMP effectively uses each feature template to perform a linear regression to fit the gradient of the loss function. This attempts to find the correlation of each feature subset with the residual of the model. It then adds the feature template that best fits this gradient, and retrains the model. The main weakness of this method is that it fits the gradient of the training error which can rapidly overfit for sparse, high-dimensional data. Ultimately, we would prefer to use a development set for feature selection.

\subsubsection{Wrapper Method}

The \emph{wrapper method} \cite{kohavi1997wrappers} is a meta-algorithm for feature selection, usually based on a validation set. We employ it in a stagewise approach to learning a sequence of templates. Given an ordering of the initial
sub-sequence and a learning procedure, we add each remaining template
to our ordering and estimate parameters, selecting as the next
template the one that gives the highest increase in development set
performance.  We begin the procedure with no templates, and repeat
the procedure until we have a total ordering over the set of feature
templates. When learning the ordering we use the same hyperparameters as will be used during final training.

While simpler than the Lasso and Matching Pursuit approaches, we empirically found this approach to outperform the others, due to the necessity of using a development set to select features for our high-dimensional application areas. 

\section{Related Work}

Our work is primarily inspired by previous research on cascades of
classifiers; however, it differs significantly by approximating the
score of a single linear model---scoring as few of its features as
possible to obtain sufficient confidence.

We pose and address the question of whether a single, interacting set
of parameters can be learned such that they efficiently both (1)
provide high accuracy and (2) good confidence estimates throughout
their use in the lengthening prefixes of the feature template
sequence.  (These two requirements are both incorporated into our
novel parameter estimation algorithm.)  In contrast, other work
\cite{weiss-taskar-13,he-et-al-13} learns a separate classifier to
determine when to add features.  Such heavier-weight approaches are
unsuitable for our setting, where the core classifier's features and
scoring are already so cheap that adding complex decision-making would
cause too much computational overhead.

Other previous work on cascades uses a series of increasingly complex
models, such as the Viola-Jones face detection cascade of
classifiers \shortcite{viola2001rapid}, which applies boosted trees
trained on subsets of features in increasing order of complexity as
needed, aiming to reject many sub-image windows early in
processing.  We allow scores from each layer to directly
affect the final prediction, avoiding duplicate incorporation of
evidence. 

Our work is also related to the field of learning and inference under
test-time budget constraints \cite{grubb-bagnell-12,trapznikov-saligrama-13}. However, common approaches to this problem
also employ auxiliary models to rank which feature to add next, and
are generally suited for problems where features are expensive to
compute ({\it e.g} vision) and the extra computation of an auxiliary
pruning-decision model is offset by substantial reduction in feature
computations \cite{weiss-taskar-13}. Our method uses confidence scores
directly from the model, and so requires no additional computation,
making it suitable for speeding up classifier-based NLP methods that
are already very fast and have relatively cheap features.

Some cascaded approaches strive at each stage to prune the number
of possible output structures under consideration, whereas in our case
we focus on pruning the input features. For example, \newcite{xu-et-al-13} learn 
a tree of classifiers that sub-divides the set of classes to minimize average test-time cost.
\newcite{chen-et-al-12} similarly use a linear cascade instead
of a tree.  \newcite{weiss-taskar-10} prune output labels in the
context of structured prediction through a cascade of increasingly
complex models, and \newcite{rush2012vine} successfully apply these structured prediction cascades to the task of graph-based dependency parsing.

In the context of NLP, \newcite{he-et-al-13} describe a method for
dynamic feature template selection at test time in graph-based
dependency parsing.  Their technique is particular to the parsing
task---making a binary decision about whether to lock in edges in the
dependency graph at each stage, and enforcing parsing-specific,
hard-coded constraints on valid subsequent edges.  Furthermore, as
described above, they employ an auxiliary model to select features.

\newcite{he2012cost} share our goal to speed test time prediction by dynamically selecting features, but they also learn an additional model on top of a fixed base model, rather than using the training objective of the model itself.

While our comparisons above focus on other methods of \emph{dynamic} feature selection, there also exists related work in the field of general (static) feature selection. The most relevant results come from the applications of \emph{group sparsity}, such as the work of \newcite{martins2011structured} in \emph{Group Lasso} for NLP problems. The Group Lasso regularizer \cite{yuan2006model} sparsifies groups of feature weights (e.g. feature templates), and has been used to speed up test-time prediction by removing entire templates from the model. The key difference between this work and ours is that we select our templates based on the test-time difficulty of the inference problem, while the Group Lasso must do so at train time. In Appendix \ref{ordering-experiments}, we compare against Group Lasso and show improvements in accuracy and speed.

Note that non-grouped approaches to selecting sparse feature subsets, such as boosting and $\ell_1$ regularization, do not achieve our goal of fast test-time prediction in NLP models, as they would not zero-out entire templates, and still require the computation of a feature for every template for every test instance.

\section{Experimental Results}

We present experiments on three NLP tasks for which greedy sequence labeling has been a successful solution: part-of-speech tagging, transition-based dependency parsing and named entity recognition. In all cases our method achieves multiplicative speedups at test time with little loss in accuracy.

\subsection{Part-of-speech tagging}

\begin{table}
\begin{center}
\begin{tabular}{|lllll|}
\hline
\bf Model/$m$ & \bf Tok. & \bf Unk. & \bf Feat. & \bf Speed \\\hline\hline
Baseline & 97.22 & 88.63 & 46 & 1x\\ 
Stagewise & 96.54 & 83.63 & 9.50 & 2.74\\
\hline

Fixed & 89.88 & 56.25 & 1 & 16.16x\\ 
Fixed & 94.66 & 60.59 & 3 & 9.54x\\
Fixed & 96.16 & 87.09 & 5 & 7.02x\\ 
Fixed & 96.88 & 88.81 & 10 & 3.82x\\ \hline

Dynamic/15 & 96.09 & 83.12 & 1.92 & {\bf 10.36x}\\ 
Dynamic/35 & 97.02 & 88.26 & 4.33 & {\bf 5.22x}\\ 
Dynamic/45 & 97.16 & 88.84 & 5.87 & 3.97x\\ 
Dynamic/50 & {\bf 97.21} & 88.95 & 6.89 & 3.41x\\  \hline

\end{tabular}
\caption{Comparison of our models using different margins $m$, with speeds measured relative to the baseline. We train a model as accurate as the baseline while tagging 3.4x tokens/sec, and in another model maintain $>97\%$ accuracy while tagging 5.2x, and $>96\%$ accuracy with a speedup of 10.3x.\label{pos-compare-table}}
\vspace{-0.5cm}
\end{center}
\end{table}

We conduct our experiments on classifier-based greedy part-of-speech tagging. Our baseline tagger uses the same features
described in \newcite{choi-palmer-12}. We evaluate our models on the Penn Treebank 
WSJ corpus \cite{marcus-et-al-93}, employing the typical split of sections used for part-of-speech tagging: 0-18 train, 19-21 development, 22-24 test. 
The parameters of our models are learned using AdaGrad \cite{Duchi2011} with $\ell_2$
regularization via regularized dual averaging \cite{Xiao2009}, and we used random search on the development set to select hyperparameters.

This baseline model ({\bf baseline}) tags at a rate of approximately 23,000 tokens per
second on a 2010 2.1GHz AMD Opteron machine with accuracy comparable to 
similar taggers \cite{Gimenez-Marquez-2004,choi-palmer-12,toutanova-et-al-03}.
On the same machine the greedy Stanford CoreNLP left3words
part-of-speech tagger also tags at approximately 23,000 tokens per
second. Significantly higher absolute speeds for all methods can be
attained on more modern machines.

We include additional baselines that
divide the features into templates, but train the templates'
parameters more simply than our algorithm. The {\bf stagewise} baseline
learns the model parameters
for each of the templates in order, starting with only one template---once each
template has been trained for a fixed number of iterations, that template's
parameters are fixed and we add the next one.
We also create a separately-trained baseline model for each fixed prefix of the feature templates ({\bf fixed}). This shows that our speedups are not simply due to
superfluous features in the later templates.

\begin{figure*}
\centering
\begin{subfigure}[b]{0.48\textwidth}
\includegraphics[width=\textwidth]{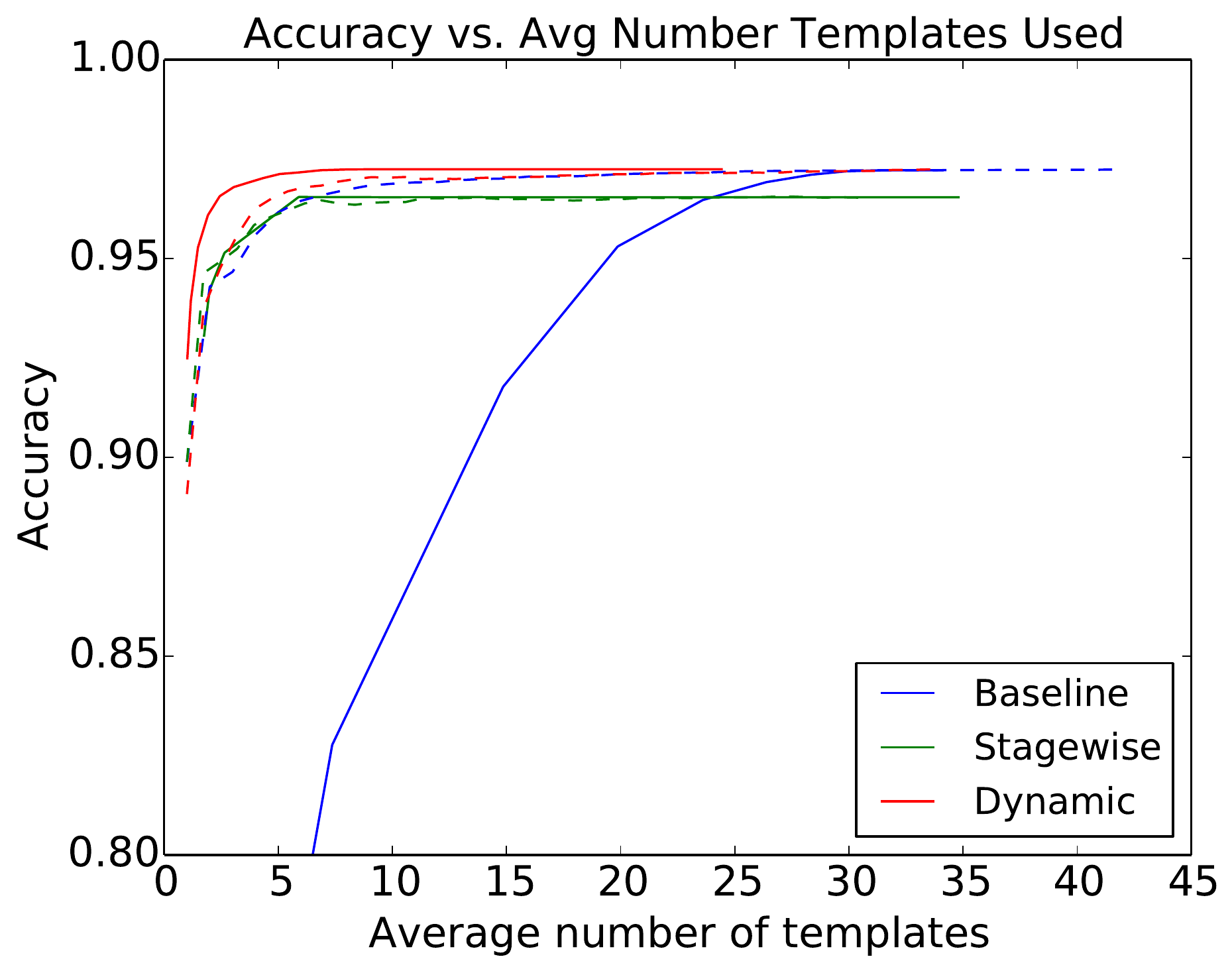}
\end{subfigure}~
\begin{subfigure}[b]{0.5\textwidth}
\includegraphics[width=\textwidth]{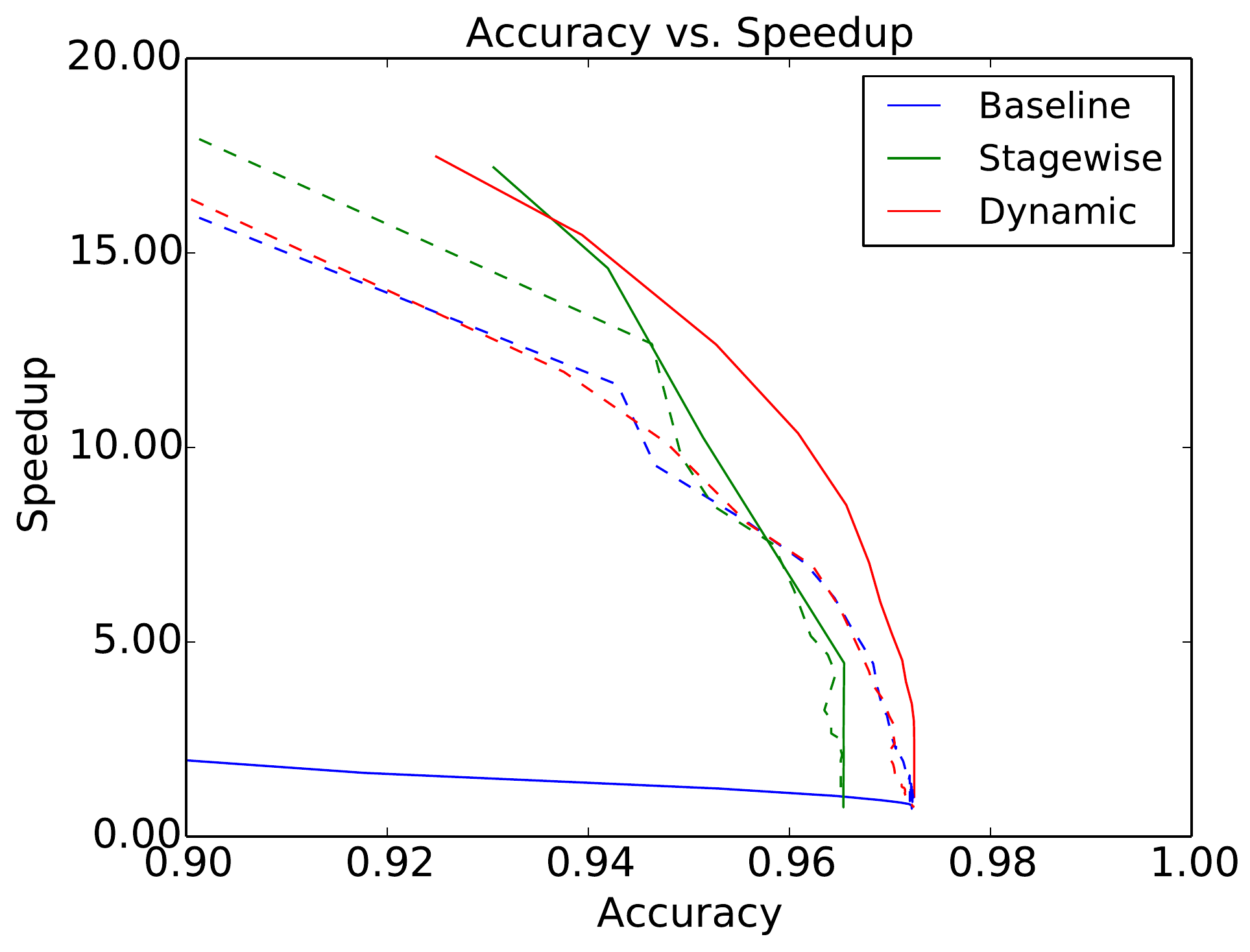}
\end{subfigure}
\caption{Left-hand plot depicts test accuracy as a function of the average number of templates used to predict. Right-hand plot shows speedup as a function of accuracy. Our model consistently achieves higher accuracy while using fewer templates resulting in the best ratio of speed to accuracy.}\label{fig:plots}
\end{figure*}

Our main results are shown in Table~\ref{pos-compare-table}.
We increase the speed of our baseline POS tagger by a factor of 5.2x
without falling below 97\% test accuracy.  By tuning our training
method to more aggressively prune templates, we achieve speed-ups of over 10x
while providing accuracy higher than 96\%. It is worth noting that the results for our method ({\bf dynamic}) are all obtained from a single trained model (with hyperparameters optimized for $m=50$, which we observed gave a good speedup with nearly no lossin accuracy on the development set), the only difference being that we varied the margin at test time. Superior results for $m\neq50$ could likely be obtained by optimizing hyperparameters for the desired margin.

Results show our method ({\bf dynamic}) learns to dynamically select
the number of templates, often
using only a small fraction.
The majority of test tokens can be tagged using
only the first few templates: just over 40\% use one template, 
and 75\% require at most four
templates, while maintaining 97.17\% accuracy.
On average 6.71 out of 46 templates are used, though a small set of complicated
instances never surpass the margin and use all 46 templates. The right hand plot of Figure \ref{fig:plots} shows speedup vs.\ accuracy for various settings of the confidence margin $m$.

The left plot in Figure \ref{fig:plots}
depicts accuracy as a function of the number of templates used at
test time. We present results for both varying the number of templates directly (dashed) and margin (solid). The baseline model trained on all
templates performs very poorly when using margin-based inference, since its training objective does not learn to predict with only prefixes. When predicting using a fixed subset of templates, we use a different baseline model for each one of the 46 total template prefixes, learned with only those features; we then compare the test accuracy of
our dynamic model using template prefix $i$ to the baseline model trained on the fixed prefix $i$.
Our model performs just as well as these separately trained models,
demonstrating that our objective learns weights that allow
each prefix to act as its own high-quality classifier.

\subsubsection{Learning the template ordering}

As described in Section \ref{ordering-section}, we experimented on part-of-speech tagging with three different algorithms for learning an ordering of feature templates: Group Lasso, Group Orthogonal Matching Pursuit (GOMP), and the wrapper method. For the case of Group Lasso, this corresponds to the experimental setup used when evaluating Group Lasso for NLP in \newcite{martins2011structured}. As detailed in the part-of-speech tagging experiments of Appendix \ref{ordering-experiments}, we found the wrapper method to work best in our dynamic prediction setting. Therefore, we use it in our remaining experiments in parsing and named entity recognition. Essentially, the Group Lasso picks small templates too early in the ordering by penalizing template norms, and GOMP picks large templates too early by overfitting the train error.

\subsection{Transition-based dependency parsing}

We base our parsing experiments on the greedy, non-projective transition-based dependency parser described in \newcite{Choi2011}. Our model uses a total of 60 feature templates based mainly on the word form, POS tag, lemma and assigned head label of current and previous input and stack tokens, and parses about 300 sentences/second on a modest 2.1GHz AMD Opteron machine.

We train our parser on the English Penn TreeBank, learning the parameters using AdaGrad and the parsing split, training on sections 2--21, testing on section 23 and using section 22 for development and the Stanford dependency framework \cite{deMarneffe2008}. POS tags were automatically generated via 10-way jackknifing using the baseline POS model described in the previous section, trained with AdaGrad using $\ell_2$ regularization, with parameters tuned on the development set to achieve 97.22 accuracy on WSJ sections 22-24. Lemmas were automatically generated using the ClearNLP morphological analyzer. We measure accuracy using labeled and unlabeled attachment scores excluding punctuation, achieving a labeled score of 90.31 and unlabeled score of 91.83, which are comparable to similar greedy parsers \cite{Choi2011,Honnibal2013}.

Our experimental setup is the same as for part-of-speech tagging. We compare our model ({\bf dynamic}) to both a single {\bf baseline} model trained on all features, and a set of 60 models each trained on a prefix of feature templates. Our experiments vary the margin used during prediction (solid) as well as the number of templates used (dashed).

As in part-of-speech tagging, we observe significant test-time speedups when applying our method of dynamic feature selection to dependency parsing. With a loss of only 0.04 labeled attachment score (LAS), our model produces parses 2.7 times faster than the baseline. As listed in Table \ref{parse-compare-table}, with a more aggressive margin our model can parse more than 3 times faster while remaining above 90\% LAS, and more than 5 times faster while maintaining accuracy above 88.5\%.


\begin{figure}
\centering
\includegraphics[width=0.5\textwidth]{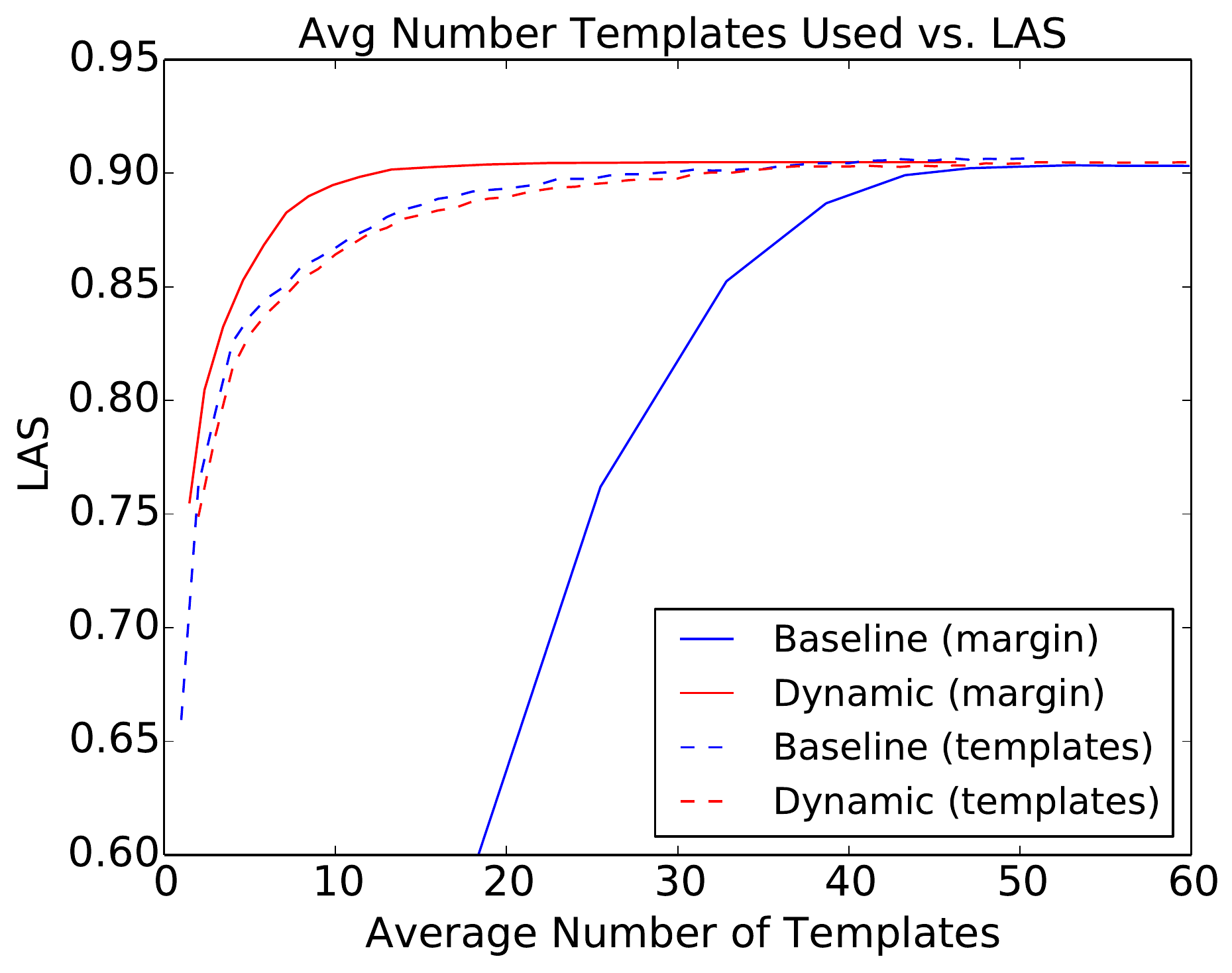}
\caption{Parsing speedup as a function of accuracy. Our model achieves the highest accuracy while using the fewest feature templates. \label{parse-templates-accuracy}}
\end{figure}

In Figure \ref{parse-templates-accuracy} we see not only that our {\bf dynamic} model consistently achieves higher accuracy while using fewer templates, but also that our model ({\bf dynamic}, dashed) performs exactly as well as separate models trained on each prefix of templates ({\bf baseline}, dashed), demonstrating again that our training objective is successful in learning a single model that can predict as well as possible using any prefix of feature templates while successfully selecting which of these prefixes to use on a per-example basis. 

\begin{table}
\centering
\begin{tabular}{|lllll|}
\hline
\bf Model/$m$ & \bf LAS & \bf UAS & \bf Feat. & \bf Speed \\ \hline\hline
Baseline & 90.31 & 91.83 & 60 & 1x \\ \hline
Fixed & 65.99 & 70.78 & 1 & 27.5x \\ 
Fixed & 86.87 & 88.81 & 10 & 5.51x \\ 
Fixed & 88.76 & 90.51 & 20 & 2.83x \\ 
Fixed & 89.04 & 90.71 & 30 & 1.87x \\ \hline
Dynamic/6.5 & 88.63 & 90.36 & 7.81 & 5.16x \\ 
Dynamic/7.1 & 89.07 & 90.73 & 8.57 & 4.66x \\ 
Dynamic/10 & 90.16 & 91.70 & 13.27 & 3.17x \\ 
Dynamic/11 & 90.27 & 91.80 & 15.83 & 2.71x \\ \hline
\end{tabular}
\caption{Comparison of our baseline and templated models using varying margins $m$ and numbers of templates. \label{parse-compare-table}}
\end{table}


\subsection{Named entity recognition}
We implement a greedy left-to-right named entity recognizer based on \newcite{ratinov2009design} using a total of 46 feature templates, including surface features such as lemma and capitalization, gazetteer look-ups, and each token's \emph{extended prediction history}, as described in \cite{ratinov2009design}. Training, tuning, and evaluation are performed on the CoNLL 2003 English data set with the BILOU encoding to denote label spans.

Our baseline model achieves F1 scores of $88.35$ and $93.37$ on the test and development sets, respectively, and tags at a rate of approximately $5300$ tokens per second on the hardware described in the experiments above. We achieve a $2.3$x speedup while maintaining F1 score above $88$ on the test set.

\begin{table}
\centering
\begin{tabular}{|llll|}
\hline
\bf Model/$m$ & \bf Test F1 & \bf Feat. & \bf Speed \\ \hline\hline
Baseline & 88.35 & 46 & 1x \\ \hline
Fixed &  65.05 & 1 & 19.08x \\ 
Fixed & 85.00 & 10 & 2.14x \\
Fixed & 85.81 & 13 & 1.87x \\ \hline
Dynamic/3.0 & 87.62 & 7.23 & 2.59x \\
Dynamic/4.0 & 88.20 & 9.45 & 2.32x \\
Dynamic/5.0 & 88.23 & 12.96 & 1.96x \\ \hline

\end{tabular}
\caption{Comparison of our baseline and templated NER models using varying margin $m$ and number of templates.
 \label{ner-compare-table}}
\end{table}

\section{Conclusions and Future Work}

By learning to dynamically select the most predictive features at test time, our algorithm provides significant speed improvements to classifier-based structured prediction algorithms, which themselves already comprise the fastest methods in NLP. Further, these speed gains come at very little extra implementation cost and can easily be combined with existing state-of-the-art systems. Future work will remove the fixed ordering for feature templates, and dynamically add additional features based on the current scores of different labels.

\section{Acknowledgements}
This work was supported in part by the Center for Intelligent Information Retrieval, in part by
DARPA under agreement number FA8750-13-2-0020, and in part by NSF grant \#CNS-0958392.
The U.S.\ Government is authorized to reproduce and distribute reprint for Governmental purposes
notwithstanding any copyright annotation thereon. Any opinions, findings and conclusions or recommendations expressed in this material are those of the authors and do not necessarily reflect those of the sponsor.

\bibliographystyle{acl}
\bibliography{paper}

\newpage
\onecolumn
\appendix
\begin{center}
\Large{Supplementary Material}
\end{center}

\section{Experiments: Learning Template Ordering}
\label{ordering-experiments}

As described in Section \ref{ordering-section}, we experimented with 3 different algorithms for learning an ordering of feature templates: Group lasso ({\bf Lasso}), which prunes feature templates based on their norm after $\ell_1$ regularization; Group orthogonal matching pursuit ({\bf GOMP}), which selects features by iteratively training a model using the existing features then adds the template that maximizes Eqn. \ref{eq:regularized-gomp}; and the wrapper method ({\bf wrapper}). We use the methods of setting regularization parameters for Lasso and GOMP discussed in Section \ref{ordering-section}. Note that for the case of Group Lasso, this corresponds to the experimental setup used when evaluating Group Lasso for NLP in \newcite{martins2011structured}.

\begin{table}
\centering
\begin{tabular}{|llll|}
\hline
\bf Model & \bf Templates & \bf Accuracy & \bf Speed \\ \hline\hline
Baseline & 46 & 97.22 & 1x \\ \hline 
Lasso & 6 & 90.53 & 10.97x \\
Lasso & 10 & 94.33 & 7.67x \\
Lasso & 15 & 94.84 & 5.23x \\
Lasso & 23 & 96.19 & 3.31x \\ \hline
GOMP & 6 & 92.18 & 6.83x \\
GOMP & 10 & 94.15 & 4.29x \\
GOMP & 15 & 96.46 & 2.83x \\
GOMP & 23 & 96.81 & 1.96x \\ \hline
Wrapper & 6 & 96.45 & 6.13x \\
Wrapper & 10 & 96.88 & 3.82x \\
Wrapper & 15 & 97.01 & 2.62x \\
Wrapper & 23 & 97.15 & 1.70x \\ \hline
\end{tabular}
\caption{Comparison of the wrapper method for learning template orderings with group lasso and group orthogonal matching pursuit. Speeds are measured relative to the baseline, which is about 23,000 tokens/second on a 2.1GHz AMD Opteron machine. \label{ordering-table}}
\end{table}

We compare the different methods for picking template orderings in Table \ref{ordering-table}, training and testing models for WSJ POS tagging. To keep concerns separate, we use standard sequential prediction using all templates rather than our dynamic prediction method. We found the wrapper method to be the most successful towards our goal of achieving high accuracy while using as few templates as possible, which is important when using dynamic prediction. While the wrapper method comes within 0.15\% of state-of-the-art accuracy using the first 23 out of 46 total templates, neither GOMP nor Lasso are able to exceed 97\% accuracy with so few templates.

In terms of speed, Lasso outperforms both GOMP and the wrapper method, predicting 3 times as fast with 23 templates as the baseline, whereas GOMP predicts twice as fast, and the wrapper method 1.7 times baseline speed.
It is initially puzzling that for a given number of templates, each model does not achieve the same speed increase. Since each template has only a single active feature for a given test instance, we are computing the same number of multiplications and additions for each model. However, the different models are selecting very different feature templates, whose features can take a widely different amount of time to compute. For example, creating and hashing the string representing a trigram conjunction into a sparse vector takes much longer than creating a feature whose template has only two possible values, \emph{true} and \emph{false}. 

This behavior is a reflection of some interesting properties of the template selection methods, which help to explain the superior performance of the wrapper method. 

Since the Group Lasso penalizes the norms of the templates as a surrogate for the sparsity (a 0-1 loss), it is naturally biased against large templates and will always include very small templates -- the early stages of the regularization path will contain these small templates whose features can be quickly computed. Another likely source of speedup arises from the impact on cache locality of using much smaller cardinality weight vectors. This also explains the poor performance of the induced template ordering. Including small templates early on runs at cross-purposes to our goal of placing highly predictive templates up front for our dynamic prediction algorithm.

In contrast, the wrapper method and GOMP both pick larger, more predictive templates early on since they attempt to minimize loss functions that are unrelated to template size. While GOMP produces better orderings than lasso, in this case the wrapper method works better because the template ordering produced by GOMP is unable to incorporate signal from a validation set and overfits the training set. 

We find that the wrapper method works well with our dynamic training objective, which benefits from predictive, though expensive, features early on in the ordering. When using dynamic prediction, the speed per template used is offset by using significantly fewer templates on examples that are easier to classify.

\section{Sparse Regularized Group Orthogonal Matching Pursuit}
\label{gomp-derivation}

Group Orthogonal Matching Pursuit (GOMP) picks a stagewise ordering of feature templates to add to a generalized linear model. At each stage, GOMP effectively uses each feature template to perform a linear regression to fit the gradient of the loss function. This attempts to find the correlation of each feature subset with the residual of the model. It then adds the feature template that best fits this gradient, and retrains the model. We adapt this algorithm to the setting of high-dimensional NLP problems by efficiently inverting the covariance matrices of the feature templates, and regularizing the computation of the residual correlation. This results in a scalable feature selection technique for our problem setting, detailed below.

For purposes of exposition, we will break from the notation of Section \ref{linear-model-section} that combines features of $x$ and $y$ since the algorithm is designed from a linear-algebraic, regression standpoint that considers the design matrix $X$ and the label matrix $Y$ as separate entities. We will call each group of features $G_i$, its associated design matrix $X_i$ (the feature matrix for that template).

At each step $k$, we use our selected set of feature templates/groups and compute the gradient of our loss function on the training data set, call it $r^{(k)}$, and then we select a new feature template $G_i$ with corresponding design matrix $X_i$ to add to our selected groups, by finding the index that maximizes:
\begin{align}
\label{eq:normal-gomp}
\argmax_i \textbf{tr} \big{(} (r^{(k)})^\top X_i (X_i^\top X_i)^{-1} X_i^\top r^{(k)} \big{)}
\end{align}
After adding this template, we retrain the model on the selected set of templates and repeat. Note that this appears to be a difficult optimization problem to solve: some of our templates have hundreds of thousands or even millions of features and computing the inverse $(X^\top_i X_i)^{-1}$ could be expensive -- however, due to the special structure of our NLP problems, where each feature template contains one-hot features, this covariance matrix is diagonal and hence trivially invertible.

However, we find in practice that due to the large number of features and relatively small number of examples in our NLP models, this picks very-high cardinality feature templates early on that generalize poorly. The reason becomes apparent when we notice that the correlation-finding subroutine, Equation \eqref{eq:normal-gomp}, is essentially an un-regularized least squares problem, attempting to regress the loss function gradient onto the data matrices for each template. This suggests we should try a form of regularization, by using some template-dependent constant $\alpha_i$ to regularize the inversion of the covariance matrix:
\begin{align}
\label{eq:regularized-gomp}
\argmax_i \textbf{tr} \big{(} (r^{(k)})^\top X_i (X_i^\top X_i + \lambda_i I)^{-1} X_i^\top r^{(k)} \big{)}
\end{align}
Heuristically, we pick this regularization parameter to be a low fractional power of the dimension size for each feature template $\lambda_i = \textbf{col}(X_i)^{\alpha_i}$, where $\alpha_i$ is picked to be in the regime of $[0.25, 0.5]$.

\end{document}